\documentclass[final]{cvpr}

\usepackage{times}
\usepackage{epsfig}
\usepackage{graphicx}
\usepackage{amsmath}
\usepackage{amssymb}

\usepackage{enumitem}
\usepackage[ruled,vlined]{algorithm2e}
\usepackage{amsthm}
\usepackage{subfigure}
\usepackage{multirow}
\usepackage[table,xcdraw]{xcolor}

\usepackage[pagebackref=true,breaklinks=true,colorlinks,bookmarks=false]{hyperref}

\def\cvprPaperID{3343} 
\def\confYear{CVPR 2021}

\begin{document}

\title{Fast Bayesian Uncertainty Estimation and Reduction of Batch Normalized Single Image Super-Resolution Network}

\author{Aupendu Kar, Prabir Kumar Biswas\\
Indian Institute of Technology Kharagpur, WB, India\\
Webpage: \href{https://aupendu.github.io/sr-uncertainty}{aupendu.github.io/sr-uncertainty}\\
{\tt\small mailtoaupendu@gmail.com, pkb@ece.iitkgp.ac.in}
}

\maketitle

\begin{abstract}
Convolutional neural network (CNN) has achieved unprecedented success in image super-resolution tasks in recent years. However, the network's performance depends on the distribution of the training sets and degrades on out-of-distribution samples. This paper adopts a Bayesian approach for estimating uncertainty associated with output and applies it in a deep image super-resolution model to address the concern mentioned above. We use the uncertainty estimation technique using the batch-normalization layer, where stochasticity of the batch mean and variance generate Monte-Carlo (MC) samples. The MC samples, which are nothing but different super-resolved images using different stochastic parameters, reconstruct the image, and provide a confidence or uncertainty map of the reconstruction. We propose a faster approach for MC sample generation, and it allows the variable image size during testing. Therefore, it will be useful for image reconstruction domain. Our experimental findings show that this uncertainty map strongly relates to the quality of reconstruction generated by the deep CNN model and explains its limitation. Furthermore, this paper proposes an approach to reduce the model's uncertainty for an input image, and it helps to defend the adversarial attacks on the image super-resolution model. The proposed uncertainty reduction technique also improves the performance of the model for out-of-distribution test images.  To the best of our knowledge, we are the first to propose an adversarial defense mechanism in any image reconstruction domain. 
\end{abstract}

\section{Introduction}
\label{sec:intro}

% What is Super-resolution
Single image super-resolution (SISR) is an ill-posed low vision problem, where we upscale the image to increase the image's spatial resolution. Besides improving the perceptual quality of images for human interpretation, SISR boosts other computer vision tasks' performance~\cite{srb}. Due to the advancement of deep learning techniques, the community has developed a state-of-the-art SISR network using different topology-based deep neural architectures~\cite{srs}. However, due to deep learning models' black-box nature, it is always hard to trust CNN's outcome and the limitations of the models are unknown. 

\begin{figure}[t]
    \centering
    \subfigure{
        \includegraphics[height=3.1cm]{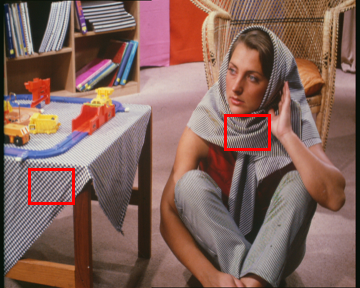}} \hspace{-0.1cm} 
    \subfigure{
        \includegraphics[height=3.1cm]{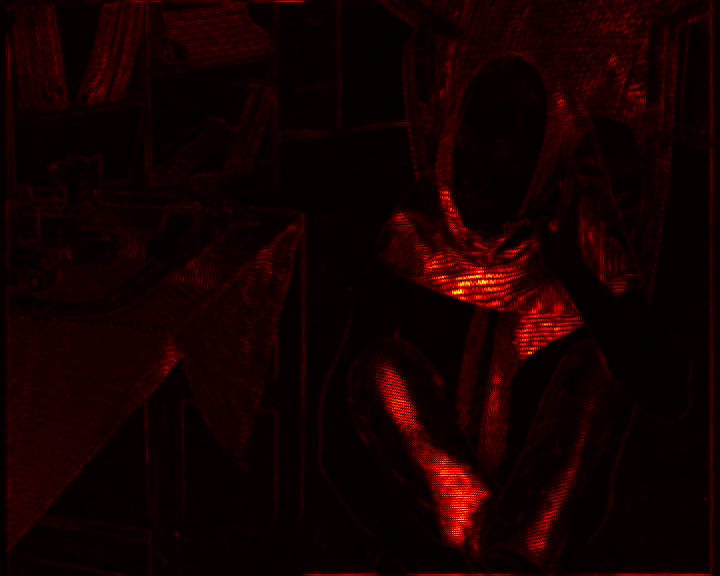}} \hspace{-0.2cm} \vspace{-0.3cm}
    \subfigure{
        \includegraphics[height=3.1cm]{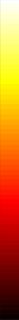}} \hspace{-0.2cm}
    \subfigure{
        \includegraphics[width=2cm]{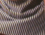}} \hspace{-0.2cm}
    \subfigure{
        \includegraphics[width=2cm]{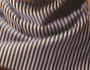}} \hspace{-0.2cm}
    \subfigure{
        \includegraphics[width=2cm]{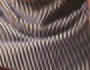}} \hspace{-0.2cm} \vspace{-0.3cm}
    \subfigure{
        \includegraphics[width=2cm]{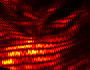}} 
        \hspace{-0.2cm}
    \setcounter{subfigure}{0}
    \subfigure[LR Image]{
        \includegraphics[width=2cm]{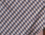}} \hspace{-0.2cm}
    \subfigure[HR Image]{
        \includegraphics[width=2cm]{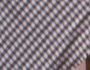}} \hspace{-0.2cm}
    \subfigure[SR Image]{
        \includegraphics[width=2cm]{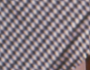}} \hspace{-0.2cm}
    \subfigure[Uncertainty]{
        \includegraphics[width=2cm]{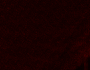}}
    \caption{In the first row images, the left image is an LR image, and the model's uncertainty during upscaling is in the right. Two cropped regions have different textures. We can observe from two patches in the second and third-row that whenever the model fails to reconstruct the texture correctly (second-row image), it leads to higher uncertainty.}
    \label{fig: GA}
\end{figure}
%\vspace{-1cm}
% What is uncertainty and it's requirement in SISR
Understanding the model's limitations is a crucial part of many machine learning systems. Deep learning models learn powerful abstract representations from high-dimensional images to map it to the outputs. The output results on the unknown test data are often considered blindly and believed to be reliable, which is not always true. Bayesian uncertainty plays a vital role in determining the confidence of the model during testing. Uncertainty is a powerful tool for any prediction and reconstruction system, and the confidence of the system's output helps in the decision-making process. We use the concept of uncertainty for image super-resolution. SISR techniques using deep learning (DL) models learn features from the training image set distribution. However, real-world pictures are entirely different and contain more complex textures, which may differ from the training set. Unseen low-resolution (LR) textures during test time can produce an inappropriate reconstruction. We also have witnessed that some artifacts, blurriness, or distortions in an image due to adversarial perturbation can significantly degrade the performance of DL based models~\cite{advattack, choi2019evaluating}. Some deformation in reconstructing LR facial images may lead to wrong output in a recognition system. Any deformed reconstruction in tumor image may lead to the incorrect estimation of tumor size. Therefore, uncertainty in DL-based reconstruction models improves the transparency and trustability of the reconstruction system.

%How we apply uncertainty
Bayesian approaches in super-resolution provide the posterior distribution of the reconstructed high-resolution (HR) image. Recent progress in Bayesian DL approaches uses Monte-Carlo (MC) samples that come from a posterior distribution via dropout~\cite{dropout} or batch-normalization~\cite{batchnorm}. Dropout during testing or stochastic batch mean-variance during testing helps to generate MC samples~\cite{bayesiandropout, bayesianbatchnorm}. Monte Carlo methods for deep learning model uncertainty estimation is successfully applied to classification, segmentation~\cite{bayesiansegnet, abhi}, camera relocalization~\cite{crelo} problems. In our work, we use batch-normalization uncertainty to analyze SISR model uncertainty.

% What we propose
In this article, we propose a Bayesian approach for SISR. For this purpose, we use a widely used batch-normalization layer in our SISR network to generate MC samples. Those samples are different possible super-resolved images from a single LR image. We use the mean of those images to get the reconstruction, and the standard deviation between those images gives the uncertainty of reconstruction. We also propose a faster approach for generating MC samples, where we generate MC samples in a single forward pass instead of multiple. Due to this, it is useful in real-time applications. We also show the importance of uncertainty in adversarial perturbation and non-ideal LR image premises, and we have made an effort to understand the SISR model limitations. We also introduce a signed gradient-based uncertainty reduction technique for any test image. We perform signed gradient dependent controlled perturbation on LR images and successfully defended adversarial attacks on the SISR model. We also achieved performance improvement on non-ideal LR images.

The key contributions of our work can be summarised as follows.
\begin{itemize}[noitemsep,nolistsep]
   \item We propose a faster implementation of Monte Carlo batch-normalization uncertainty to generate MC samples and overcome the hurdle of variable image size.
   \item We address the uncertainty of deep SISR models using Bayesian approach to measure it. To the best of our knowledge, we are the first to estimate uncertainty in deep learning models for image reconstruction. We also discuss the advantages and implications of uncertainty in SISR and its usefulness in understanding the model's limitations and outcome.
   \item We also propose an approach to reduce the model uncertainty for a test image. The uncertainty reduction method acts as a defense against the adversarial attack on the SISR model and proves to be beneficial for out-of-distribution noisy low-resolution images.
\end{itemize} 

\section{Prior Art}
\label{sec:survey}

\textbf{Single image super-resolution:} SISR has extensive literature due to different studies in the last few decades. Recent advancement of deep learning (DL) methods has achieved significant improvement in that field. VDSR~\cite{vdsr} proposed a deeper architecture and showed performance improves with the increase of network depth and converge faster using residual learning. After that, different DL based approaches~\cite{rdn, srgan, anwar2020densely, dai2019second, drbn, drcn} have been proposed and achieved state-of-the-art performance in the standard datasets. In our work, we used VDSR architecture for uncertainty analysis as it is the first deep architecture for SISR.

\textbf{Bayesian uncertainty:} Bayesian models are generally used to model uncertainty, and different approaches have been developed to adapt NNs to Bayesian reasoning, like placing a prior distribution over parameters. Due to difficulty in inferencing~\cite{galphd} of Bayesian neural networks (BNNs), some approaches~\cite{bayesiandropout, bayesianbatchnorm} have been taken to approximate BNNs. Bayesian deep learning approaches utilized MC samples generated via dropout~\cite{dropout} or batch-normalization (BN)~\cite{batchnorm} to approximate the posterior distribution. Dropout can be treated as an approximate Bayesian model with multiple predictions through the trained model by sampling predictions using different dropout masks, and In the case of BN, stochastic parameters batch mean and batch variance are used to generate multiple predictions. Thus, the batch-normalized neural network can be approximated to the Bayesian model~\cite{bayesianbatchnorm}. We use it for SISR as it is generally common in image reconstruction applications.

\section{Methodology}
\label{sec:method}

We introduce a Bayesian approach on SISR that produces high-resolution images and a confidence map of the reconstruction. In this regard, we discuss a brief background of Bayesian inference in this section. After that, our network architecture and its modifications are discussed for Bayesian approximation. We also present a faster approach to overcome the difficulties of estimating uncertainty in SISR applications. In the end, we introduce an uncertainty reduction technique for an unknown test image.

\subsection{Bayesian Inference}
We estimate a probabilistic function $F_{W}(I):I_{LR}\rightarrow I_{HR}$ from a training set $D=\{ \tilde{I}_{LR}, \tilde{I}_{HR} \}$ where $\tilde{I}_{LR}=\{ \tilde{I}_{LR_{1}}, ..., \tilde{I}_{LR_{n}} \}$ are LR image set and its corresponding HR image set $\tilde{I}_{HR}=\{ \tilde{I}_{HR_{1}}, ..., \tilde{I}_{HR_{n}} \}$. This function is approximated to generate  most likely high-resolution image ${I^{'}}_{HR}$ from a low-resolution test image ${I^{'}}_{LR}$. So the probabilistic estimation of HR test image is described as
\begin{equation}
\resizebox{.9\hsize}{!}{$p({I^{'}}_{HR}|{I^{'}}_{LR}, D)=\int p({I^{'}}_{HR}|{I^{'}}_{LR}, W)p(W|D)dW$}
\end{equation}
where $W$ is weight parameters of a function $F_{W}(I)$. We use variational inference to approximate Bayesian modeling. Most common approach is to learn approximate distribution of weights $q_{\theta}(w)$ by minimizing the Kullback–Leibler divergence $KL(q_{\theta}(w)\parallel p(W|D))$. This yields approximate distribution
\begin{equation}
\resizebox{.9\hsize}{!}{$q({I^{'}}_{HR}|{I^{'}}_{LR}, D)=\int p({I^{'}}_{HR}|{I^{'}}_{LR}, w)q_{\theta}(w)dw$}
\end{equation}
In a batch-normalized neural network for bayesian uncertainty estimation, model parameters are $W_{L}, \gamma_{L}, \beta_{L}, \mu_{L}, {\sigma^2}_{L}$. $\theta = \left\{W_{L}, \gamma_{L}, \beta_{L}\right\}$ is learnable model weight parameters and $w = \left\{\mu_{L}, {\sigma^2}_{L}\right\}$ are stochastic parameters which are mean and variance of each layer. $q_{\theta}(w)$ is a joint distribution of weights and stochastic parameters $w$. $w_{i}$ is mean and variance of $i^{th}$ sample. 

\subsection{Network Architecture}
In this paper, we use the very deep super-resolution (VDSR) network~\cite{vdsr} as a base architecture for an experimental purpose to analyze uncertainty. Our method is a generalized approach, and any other super-resolution network can adopt without any modifications. We have used batch-normalization (BN) for uncertainty estimation, but VDSR paper has not used batch-normalization. So we introduce it in the VDSR architecture and use this after each convolutional layer except the first and last layer. In our experiments, we achieve similar results like VDSR in our batch-normalized VDSR. We also perform same experiments on another super-resolution model and for this purpose we use SR-Resnet~\cite{ledig2017photo} architecture. All the detailed experiments on SR-Resnet are described in the supplementary.

\subsection{Bayesian VDSR for Uncertainty Estimation}
We use BN to estimate the uncertainty of the SISR network, where random batch members are selected to estimate mini-batch statistics for training. We use this stochasticity to approximate Bayesian inference, and it allows us to make a meaningful estimate of uncertainty, and it is termed as Monte-Carlo Batch Normalization (MCBN)~\cite{bayesianbatchnorm}. Generally, estimated running batch mean and batch variance are used in each BN layer during testing, but here we use stochastic batch mean and variance. We have learnable model parameters that are optimized during training, and stochastic parameters like batch mean and variances help to generate MC samples from the posterior distribution of the model. We feed-forward a test image along with different randomly sampled training batches for multiple times, and due to the stochasticity of batches, it creates various reconstructed HR images. We take the mean of those MC samples to estimate reconstruction and standard deviation for uncertainty map.

\subsection{Faster approach}
The main drawback of standard MCBN uncertainty estimation is that we need to process test images with different random batches to generate MC samples, and computation time increases exponentially with the increase of the number of samples in a single batch or spatial dimension of the batch.
\begin{algorithm}[ht]
\SetAlgoLined
\KwIn{training image set $I$} 
\KwOut{layer-wise batch mean and variance set ${\widehat{w}_{L}^T = \{ \mu_{L}^T, {\sigma^{2}}_{L}^T} \}$ of trained network. Where $L$ is layer no and $T$ is maximum number of MC samples required.} 
  \SetAlgoLined\SetArgSty{}
\textit{\textbf{Training SR Network:}}\\
\For{total iterations}{
\ForAll{batches $B$ of image set $I$}{
train()\; 
}}
\textit{\textbf{$\widehat{w}_{L}^T$ estimation:}}\\
\For{$T$ batches of $B$}{
forward pass\;
estimate $\widehat{w}_{L}^T$ \;
}
\caption{Layer-wise batch mean and variance estimation for single shot MC samples generation}
 \label{algo1}
\end{algorithm}
The main challenge is that in the case of SISR, test image size varies from thousands to millions of pixels. We can not make a larger spatial batch size during training as it takes longer computation time. While we train our model using a small patch size due to the computational constraint, we have to break larger images during testing for batch processing to keep the stochastic behavior, and it can create a patchy effect in the output images. Due to this, we propose a different approach to generate MC samples in a single batch. After training, we estimate stochastic parameters $w_{L}$ of each layer using different random training batches, as shown in Algorithm~\ref{algo1}. We use the same batch shape during training and stochastic parameter estimation. We estimate these parameters in each BN layer for a batch, and like this, we create several stochastic parameters set for different batches. We use these stochastic parameters during testing to generate MC samples. One stochastic parameter set generates one MC sample. During testing, we concatenate the same test image based on the required number of MC samples. In the BN layer, we normalize each image separately using different stochastic parameters, as shown in Algorithm~\ref{algo2}. Due to this, it produces various super-resolved /HR images as MC samples, which come from a posterior distribution learned from the training dataset.

\begin{algorithm}[ht]
\SetAlgoLined
\KwIn{test image $I_{LR}$, number of MC samples $N$, batch mean and variance of layer $L$ ${\widehat{w}_{L}^N = \{\mu_{L}^N, {\sigma^{2}}_{L}^N}\}$} 
\KwOut{image mean prediction $\hat{I}_{SR}$, predictive image uncertainty $\sigma$} 
concat $I_{LR}$ for $T$ times (${I^T}_{LR}$)\;
${I^N}_{SR} = F_{W}({I^N}_{LR}, \widehat{w}_{L}^N)$\;
$\hat{I}_{SR}=mean({I^N}_{SR})$\;
$\hat\sigma=std({I^N}_{SR})$\;
 \caption{Our MCBN algorithm for SISR}
 \label{algo2}
\end{algorithm}

\subsection{Uncertainty Reduction}

Stochastic parameters of the batch-normalization layers generate different MC samples, and deviations between those samples give an uncertainty map associated with reconstructed LR image. A higher value of uncertainties mainly appears in those regions where the model fails to achieve high confidence reconstruction from LR images. Those regions are mostly the textures that are far from the training sets. Adversarial attack searches for those out-of-distribution samples where the model fails miserably. In the SISR domain, adversarial attacks perturb the LR image and push the image from the training dataset distribution of the LR image in which the model is trained. Small and visually indistinguishable perturbations on LR images lead to the visually unrealistic and unwanted synthesized texture on the reconstructed image. Even a small amount of random Gaussian noise on the Bicubic downsampled LR image can heavily degrade the performance as the degraded test image does not follow the training set distribution. Whenever a test image or specific patch of the image is far from the training set distribution, it leads to higher uncertainty.

Therefore, to bring back the sample far from the training set distribution, which eventually leads to higher uncertainty, we propose an inverse perturbation mechanism that will push the test LR image in those directions, which will subsequently lead to lowering the uncertainty for a test image. Algorithm~\ref{algo3} shows an uncertainty reduction algorithm by perturbing a test image. There are two stages in this algorithm. In the first stage, average gradient directions are determined to lower the uncertainties. The second stage is the perturbation level selection.

\begin{algorithm}[ht]
\SetAlgoLined
\KwIn{test image $I_{LR}$, number of MC samples $N$, batch mean and variance of layer $L$ ${\widehat{w}_{L}^N = \{\mu_{L}^N, {\sigma^{2}}_{L}^N}\}$} 
\KwOut{perturbed test image $I'_{LR}$} 
  \SetAlgoLined\SetArgSty{}
  
\For{total updates on $I_{LR}$}{
\textit{\textbf{Gradient directions for $I_{LR}$:}}\\
\For{total iterations, $T$}{
Random Select $t_1$ and $t_2$ \;
${I^{t_1}}_{SR} = F_{W}(I_{LR}, \widehat{w}_{L}^{t_1})$ \;
${I^{t_2}}_{SR} = F_{W}(I_{LR}, \widehat{w}_{L}^{t_2})$ \;
$\mathfrak L = \frac{1}{CWH}\sum({I^{t_1}}_{SR} - {I^{t_2}}_{SR})^2$ \;
$\mathfrak{N} = \mathfrak{N} + sign(\triangledown\mathfrak L)$
}
\textit{\textbf{Perturbation Level Selection}}\\
\For{different levels of perturbation, $\beta$}{
$I'_{LR} = I_{LR} - \beta.\frac{\mathfrak{N}}{T}$\;
Calculate Uncertainty for $I'_{LR}$, $U(\beta)$ \;
\If{$U(\beta)>U(\beta-1)$}{
$I'_{LR} = I_{LR} - (\beta-1).\frac{\mathfrak{N}}{T}$ \;
break \;
}
}
}
\caption{Uncertainty reduction for an image}
 \label{algo3}
\end{algorithm}

\paragraph{Gradient Direction Calculation}
We use the trained model $F_{W}$ and the corresponding stochastic parameters set $\widehat{w}_{L}^N$ for uncertainty reduction. For a test image $I_{LR}$, we randomly select a pair of the stochastic parameter $(\widehat{w}_{L}^{t_1}, \widehat{w}_{L}^{t_2})$ from the parameter sets and use those to reconstruct two different super-resolved images ${I^{t_1}}_{SR}$ and ${I^{t_2}}_{SR}$. Our main objective is to reduce the difference between those two images, and it can be defined by minimizing pixel-wise mean-squared loss, as shown in Algorithm~\ref{algo3}. The negative sign gradient at each pixel $\mathfrak{N}$ of the input image is the test LR image update direction to reduce uncertainty. Instead of taking a single random pair of stochastic parameters, we consider multiple pairs for better estimation of gradient directions, which eventually reduces uncertainty.

\paragraph{Perturbation Level Selection}
After obtaining the gradient directions for a test image, the main challenge is selecting the proper amount of perturbation level, which will reduce the uncertainty. If we increase the value of $\beta$, the uncertainty will decrease, and the perturbed LR image $I'_{LR}$ will come closer to the training set distribution. However, after a certain point, due to heavy perturbation, uncertainty can start increasing again. Therefore, we apply a simple technique for the proper selection of $\beta$. We gradually increase the perturbation level on the LR image and calculate the uncertainty using Algorithm~\ref{algo2} for each increment. We stop adding perturbation whenever the uncertainty starts increasing.

\section{Results and Discussion}
\label{sec:foot}

\subsection{Training Details}

We use the DIV2K dataset~\cite{dataset, datasetpaper} for training, which contains $800$ training images and $100$ images for validation. Five standard benchmark testing datasets, namely Set5~\cite{set5}, Set14~\cite{set14}, BSD100~\cite{bsd100}, Urban100~\cite{urban100}, Manga109~\cite{manga109} are used for performance analysis. We randomly extract patches of size $64\times 64$ from each HR and bicubic interpolated LR image during training for a batch update. Each batch contains $16$ patches. We augment the patches by horizontal flip, vertical flip, and $90$-degree rotation and randomly choose each augmentation with a $50\%$ probability. Each input patch is normalized into $[0, 1]$ before feeding to the network. We train each model with the PyTorch framework for $1000$ epochs, where a single epoch constitutes $1000$ batch updates. Adam optimizer~\cite{adam} is used to update the weights. The learning rate is initialized to $10^{-4}$ and reduced to half after every $200$ iterations. We use mean-squared error to optimize model parameters. 

\subsection{Monte-Carlo Samples}

\subsubsection{Number of MC Samples}

\begin{figure}[!ht]
    \centering
    \subfigure[PSNR vs MC Samples]
    {
        \includegraphics[width=4cm]{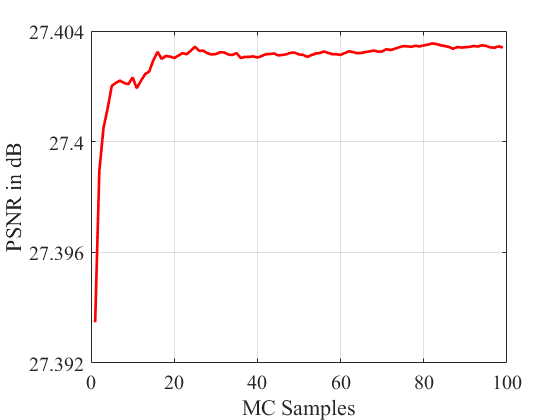}
        \label{fig: plots_PvsMC}
    } \hspace{-0.2cm} \vspace{-0.2cm}
    \subfigure[Uncertainty vs MC Samples]
    {
        \includegraphics[width=4cm]{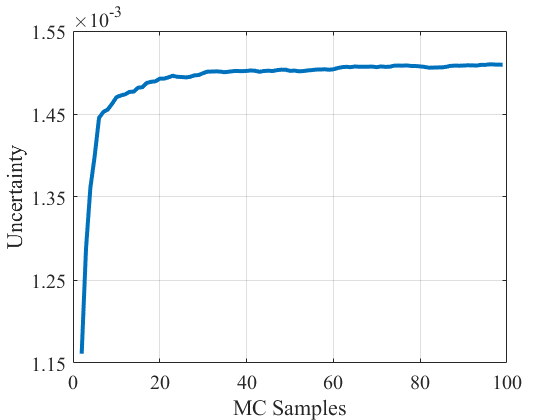}
        \label{fig: plots_UvsMC}
    } \hspace{-0.2cm}
    \subfigure[PSNR vs Uncertainty]
    {
        \includegraphics[width=4cm]{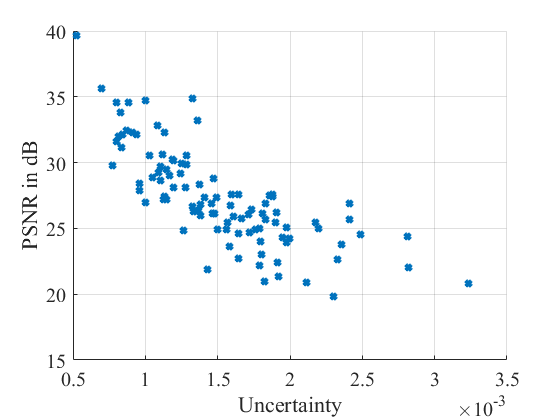}
        \label{fig: plots_PvsU}
    } \hspace{-0.2cm}
    \subfigure[Uncertainty vs Adv. Noise Level]
    {
        \includegraphics[width=4cm]{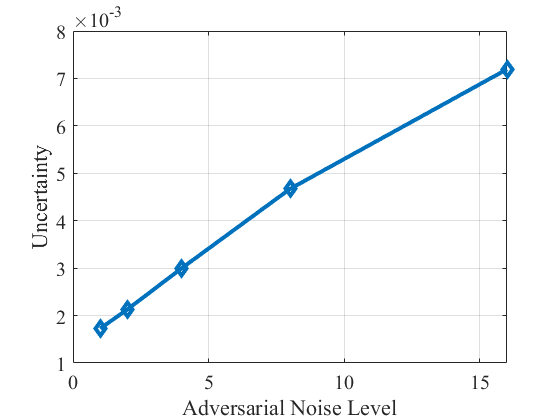}
        \label{fig: plots_UvsAdv}
    } \hspace{-0.2cm}
    \subfigure[Uncertainty vs Scale]
    {
        \includegraphics[width=4cm]{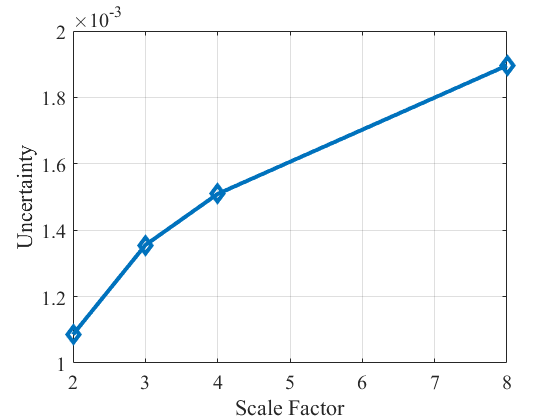}
        \label{fig: plots_UvsS}
    } \hspace{-0.2cm}
    \subfigure[Uncertainty vs Noise Level]
    {
        \includegraphics[width=4cm]{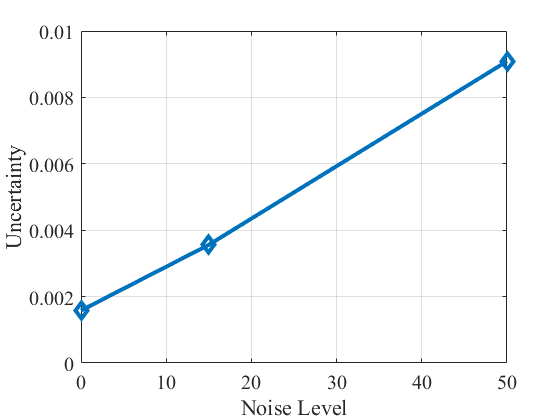}
        \label{fig: plots_UvsN}
    }
\caption{(a)-(b) present effect of PSNR and uncertainty with the increase of MC samples and (c)-(d) present impact of PSNR and adv. noise with uncertainty. (e)-(f) present the influence of Bayesian uncertainty with different Scale factor models and random noise on LR images. (zoom for the best view.)}
\label{fig: plots}
\end{figure}

We get a better estimate of uncertainty and reconstruction with the increase of MC samples, with increased inference time. Hence, a proper choice of the number of MC samples is necessary due to this trade-off. The minimum number of MC samples should produce comparable results compared to batch-normalization without stochastic mean-variance and provide a stable uncertainty estimation. Figure~\ref{fig: plots_PvsMC} shows the changes in reconstructed image quality in terms of peak signal-to-noise ratio (PSNR), and Figure~\ref{fig: plots_UvsMC} presents uncertainty associated during reconstruction with the increase in the number of MC samples. The experiments performed for those plots use all the images from the BSD100 dataset. Results show that PSNR and uncertainty increase with the increase of MC samples, and later it settles to some stable values. We observe that around 35 MC samples are sufficient to produce stable results.

\subsubsection{Fast MC Sample Generation}
\begin{table}[ht]
\centering
\begin{tabular}{|c|c|c|}
\hline
\begin{tabular}[c]{@{}c@{}}MC Samples\end{tabular} & \begin{tabular}[c]{@{}c@{}}MCBN\end{tabular} & \begin{tabular}[c]{@{}c@{}}Our approach\end{tabular} \\ \hline
5 & 14.28 &   1.0 \\ \hline
10 & 33.48 & 1.97 \\ \hline
15 & 52.67 & 2.96 \\ \hline
\end{tabular}
\caption{All values correspond to 'face' image of Set14. Five MC samples generation using our approach is used as a reference point (taken as 1.0) and other numbers represent how much more time is required to generate MC samples.}
\label{tab: fast}
\end{table}
We benchmark our faster approach against standard MCBN uncertainty estimation. The time required for generating MC samples in standard MCBN uncertainty mainly depends on the size of the image in the dataset and the number of MC samples. We overcome these two difficulties. Our approach is much faster than conventional, as shown in Table~\ref{tab: fast}. We consider $5$ MC sample generation for an image using our method as a baseline. Other values in the table exhibit how many times more GPU time is required for inference. Our approach takes $14.28$ times lesser execution time to generate $5$ samples for an image of size $276\times 276$.

\begin{figure}[!ht]
    \centering
    \subfigure
    {
        \includegraphics[height=1.7cm]{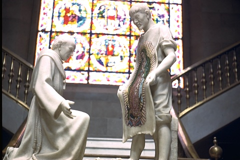}
    } \hspace{-0.2cm}
    \subfigure
    {
        \includegraphics[height=1.7cm]{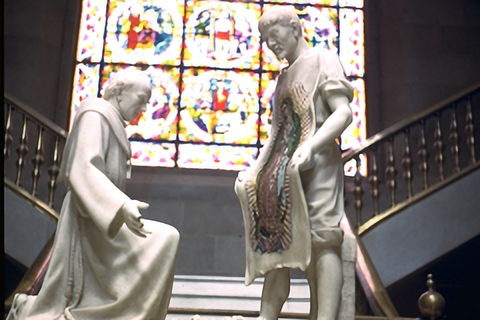}
    } \hspace{-0.2cm}
    \subfigure
    {
        \includegraphics[height=1.7cm]{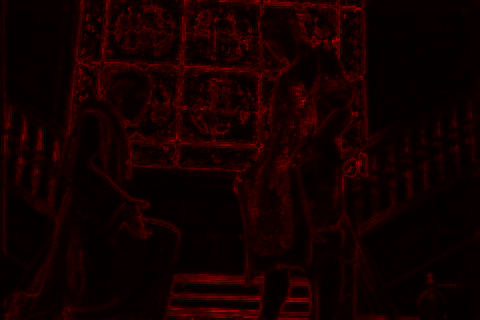}
    } \hspace{-0.3cm} \vspace{-0.3cm}
    \subfigure
    {
        \includegraphics[height=1.7cm]{fig/AttackImg/hot.PNG}
    } \hspace{-0.2cm}
    \subfigure
    {
        \includegraphics[height=1.7cm]{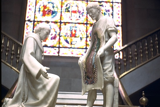}
    } \hspace{-0.2cm}
    \subfigure
    {
        \includegraphics[height=1.7cm]{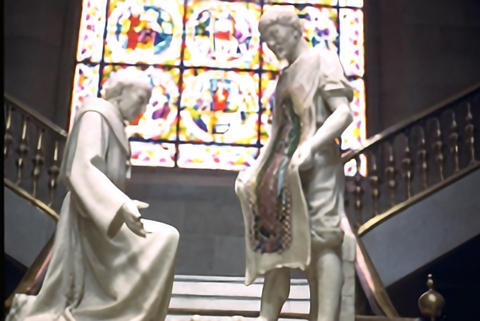}
    } \hspace{-0.2cm}
    \subfigure
    {
        \includegraphics[height=1.7cm]{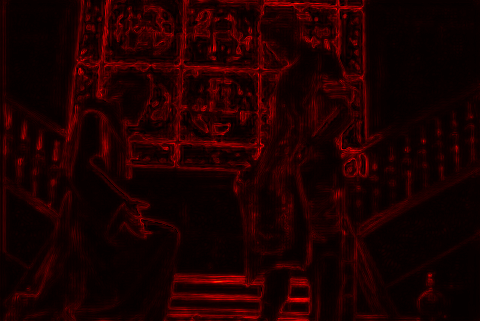}
    } \hspace{-0.3cm} \vspace{-0.3cm}
    \subfigure
    {
        \includegraphics[height=1.7cm]{fig/AttackImg/hot.PNG}
    } \hspace{-0.2cm}
    \setcounter{subfigure}{0}
    \subfigure[LR Image]
    {
        \includegraphics[height=1.7cm]{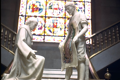}
    } \hspace{-0.2cm}
    \subfigure[SR Image]
    {
        \includegraphics[height=1.7cm]{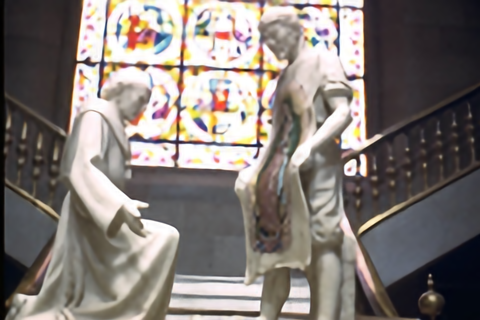}
    } \hspace{-0.2cm}
    \subfigure[Uncertainty]
    {
        \includegraphics[height=1.7cm]{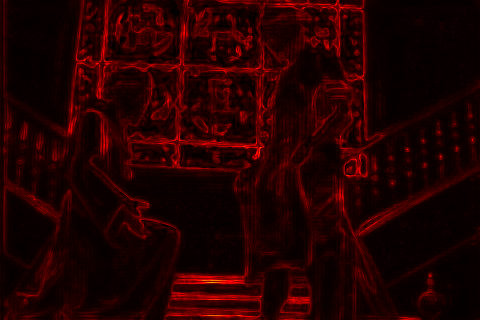}
    } \hspace{-0.3cm}
    \subfigure
    {
        \includegraphics[height=1.7cm]{fig/AttackImg/hot.PNG}
    } 
\caption{Visual representation of the reconstructed image and corresponding uncertainty map for different scale factor models. The first-row image is for scale factor $\times 2$, the second-row image is for scale factor $\times 3$, and the third-row image is for scale factor $\times 4$.}
\label{fig: scale}
\end{figure}

\subsection{Understanding Uncertainty}
\subsubsection{Behaviour of Uncertainty}

We compare the average uncertainty with the quality of reconstruction, and we use PSNR as an image quality metric. Figure~\ref{fig: plots_PvsU} shows a strong relationship between uncertainty and image quality using $100$ images of BSD100 dataset. PSNR of the images decreases with the increase of uncertainty. Figure~\ref{fig: plots_UvsS} shows that average uncertainty on BSD100 dataset increases with the increase of scale factor. However, we cannot conclude anything as we do not observe the same trend in SR-ResNet architecture as shown in supplementary. Figure~\ref{fig: scale} shows visual representations of the uncertainty for different scale factors. Uncertainty also increases with the increase of random Gaussian noise perturbation, as shown in Figure~\ref{fig: plots_UvsN}. As models are trained using bicubic downsampled noiseless images, the random noise in LR images does not match the training sets' distribution. Therefore, it leads to an increase in uncertainty. The test image moves from the ideal condition with the increase of noise in the LR image, leading to more uncertainty.

\subsubsection{Uncertainty in Adversarial attack}

Adversarial images are those strategically perturbed images where the model fails to perform. Adversarial images or patches can degrade performance significantly, and it makes deep learning models lesser trustworthy. We should know the limitations of the model and be aware of failure cases for high-risk applications like medical imaging.

\begin{figure}[!ht]
    \centering
    \subfigure
    {
        \includegraphics[height=1.7cm]{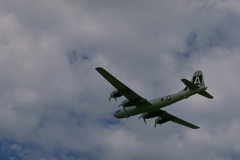}
    } \hspace{-0.3cm} 
    \subfigure
    {
        \includegraphics[height=1.7cm]{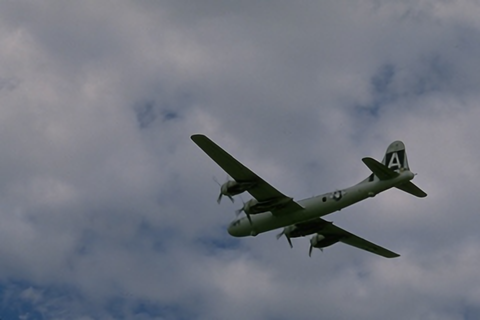}
    } \hspace{-0.3cm}
    \subfigure
    {
        \includegraphics[height=1.7cm]{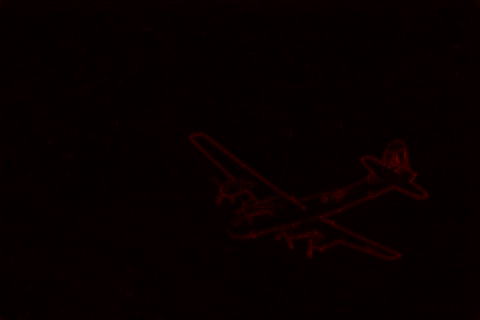}
    } \hspace{-0.3cm} \vspace{-0.3cm}
    \subfigure
    {
        \includegraphics[height=1.7cm]{fig/AttackImg/hot.PNG}
    } \hspace{-0.3cm}
    \subfigure
    {
        \includegraphics[height=1.7cm]{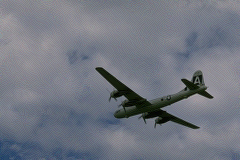}
    } \hspace{-0.3cm}
    \subfigure
    {
        \includegraphics[height=1.7cm]{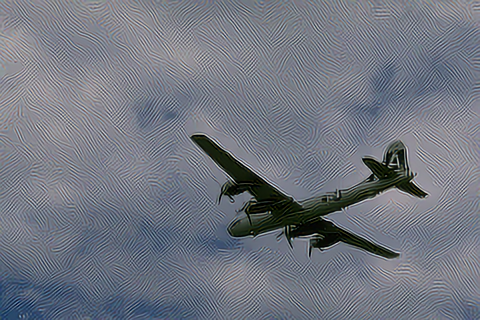}
    } \hspace{-0.3cm}
    \subfigure
    {
        \includegraphics[height=1.7cm]{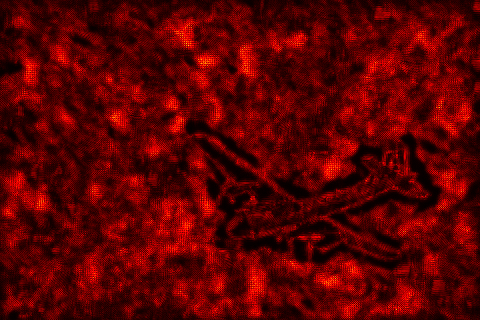}
    } \hspace{-0.3cm} \vspace{-0.3cm}
    \subfigure
    {
        \includegraphics[height=1.7cm]{fig/AttackImg/hot.png}
    } \hspace{-0.3cm}
    \setcounter{subfigure}{0}
    \subfigure[LR Image]
    {
        \includegraphics[height=1.7cm]{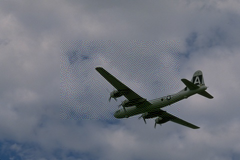}
    } \hspace{-0.3cm}
    \subfigure[SR Image]
    {
        \includegraphics[height=1.7cm]{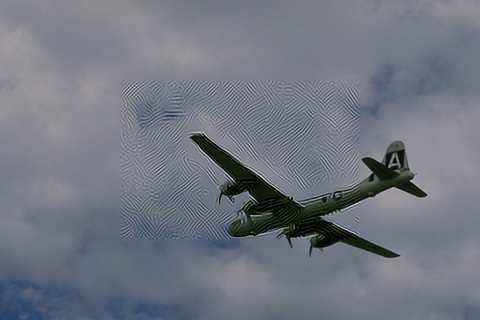}
    } \hspace{-0.3cm}
    \subfigure[Uncertainty Map]
    {
        \includegraphics[height=1.7cm]{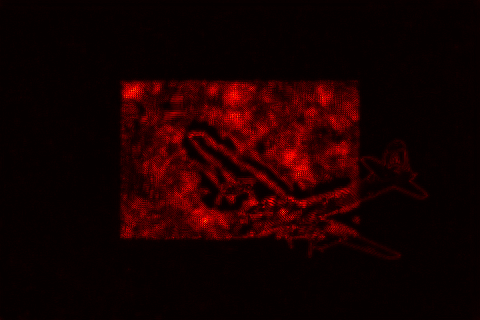}
    } \hspace{-0.3cm}
    \subfigure
    {
        \includegraphics[height=1.7cm]{fig/AttackImg/hot.png}
    }
\caption{This image is taken from the BSDS100 dataset. The first-row represents the LR image without any adversarial perturbation and its corresponding SR image and uncertainty map. The adversarial LR image and its outputs are shown in the second-row. The third-row image presents the LR image, which is partially perturbed by the adversarial attack, and the perturbed location is a square box at the center of the image.}
\label{fig: UN}
\end{figure}

The modified version of Iterative fast gradient sign method (I-FGSM) based adversarial small perturbation in the LR image can drastically degrade the model's performance and create fake details in the image~\cite{choi2019evaluating}. We analyzed the performance of the SR network under adversarial attack and found that uncertainty of the model increases with the increase of adversarial perturbation, as shown in Figure~\ref{fig: plots_UvsAdv}. Therefore, uncertainty is a crucial factor in determining black-box deep learning models' performance in any test case scenario. We generally observe uncertainty in high-frequency regions like edges of the image, which are difficult to reconstruct, as shown in the first row of Figure~\ref{fig: UN}. We can witness higher uncertainty in the airplane's edges from the first-row image without any adversarial perturbation. But, under adversarial attacks, as shown in the second row of Figure~\ref{fig: UN}, we observe higher uncertainty in the smooth regions where fake artifacts are generated. In the third row image of Figure~\ref{fig: UN}, we have shown partial adversarial attack results and the uncertainty map. We observe the partial attack region, a square box in the center of the image, produces higher uncertainty. We can conclude from here that uncertainty works on patch level. If any real-world scenario has adversarial patches in certain regions, it can be easily detected using the reconstructed image's prediction uncertainty. Hence uncertainty act as a tool to measure the capacity of the deep SISR model.

\begin{figure*}[ht]
    \centering
    \subfigure[HR]{
        \includegraphics[width=2.4cm]{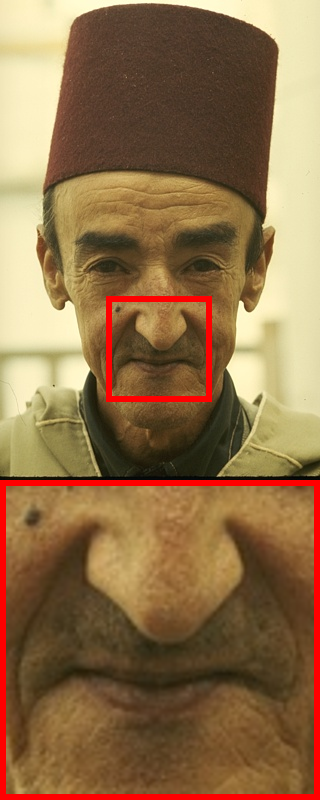}} \hspace{-0.2cm}
    \subfigure[LR]{
        \includegraphics[width=2.4cm]{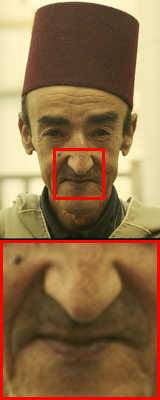}} \hspace{-0.2cm}
    \subfigure[Adversarial LR]{
        \includegraphics[width=2.4cm]{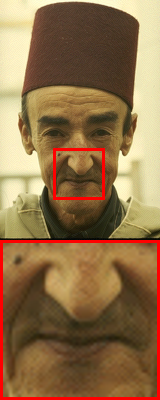}} \hspace{-0.2cm}
    \subfigure[SR (No Def.)]{
        \includegraphics[width=2.4cm]{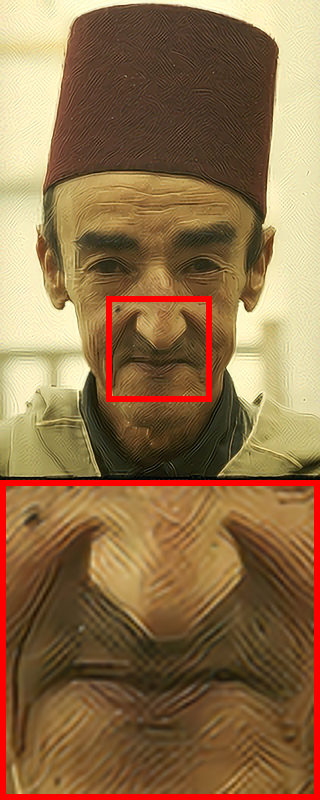}} \hspace{-0.2cm}
    \subfigure[SR (After Def.)]{
        \includegraphics[width=2.4cm]{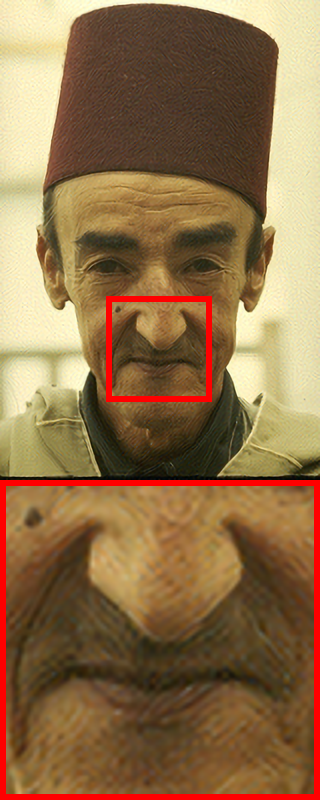}} \hspace{-0.2cm}
    \subfigure[UN (No Def.)]{
        \includegraphics[width=2.4cm]{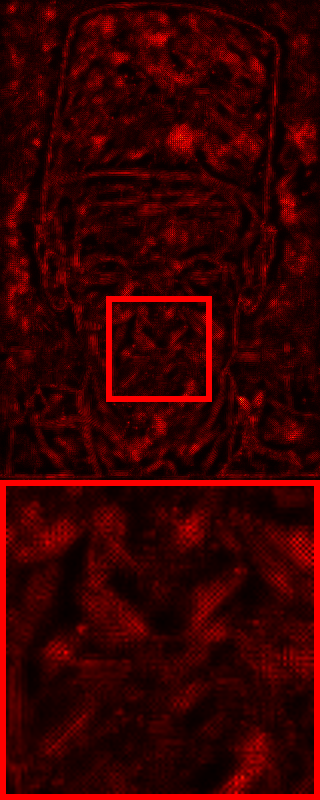}} \hspace{-0.2cm}
    \subfigure[UN (After Def.)]{
        \includegraphics[width=2.4cm]{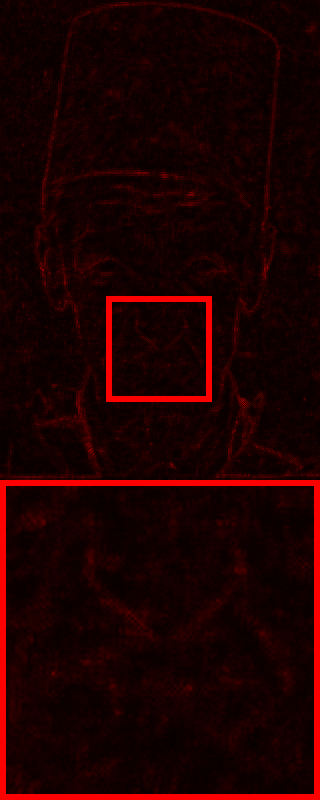}}
    \caption{Performance of our proposed adversarial defense mechanism using the Bayesian uncertainty reduction technique. The cropped region shows undesired artifacts in the SR image without any defense, and our defense mechanism successfully suppresses those artifacts. Zoom for the best view.}
    \label{fig: AdvDef}
\end{figure*}

\begin{table*}[]
\centering
\resizebox{\textwidth}{!}{%
\begin{tabular}{|c|c|c|c|c|c|}
\hline
\multirow{3}{*}{Type} & \multicolumn{5}{c|}{Scale Factor ($\times $2/ $\times $3/ $\times $4)}                                                         \\ \cline{2-6} 
                      & \multirow{2}{*}{Level of Attack} & \multicolumn{2}{c|}{PSNR}                 & \multicolumn{2}{c|}{Uncertainty}                \\ \cline{3-6} 
                      &                                  & No Defense          & After Defense       & No Defense             & After Defense          \\ \hline
No Attack               & - & 32.02/ 28.95/ 27.43 & 31.94/ 28.87/ 27.32 & 0.0011/ 0.0014/ 0.0015 & 0.0009/ 0.0011/ 0.0013 \\ \hline
\multirow{5}{*}{Attack} & 1 & 31.16/ 28.36/ 27.03 & 31.55/ 28.63/ 27.16 & 0.0014/ 0.0017/ 0.0017 & 0.0010/ 0.0013/ 0.0013 \\ \cline{2-6} 
                      & 2                                & 29.54/ 27.27/ 26.25 & 30.86/28.25/ 26.89  & 0.0019/ 0.0022/ 0.0021 & 0.0013/ 0.0014/ 0.0015 \\ \cline{2-6} 
                      & 4                                & 26.30/ 25.03/ 24.63 & 29.73/ 27.67/ 26.43 & 0.0034/ 0.0035/ 0.0030 & 0.0018/ 0.0018/ 0.0018 \\ \cline{2-6} 
                      & 8                                & 21.97/ 21.78/ 22.02 & 27.78/ 26.60/ 25.59 & 0.0058/ 0.0055/ 0.0047 & 0.0026/ 0.0024/ 0.0023 \\ \cline{2-6} 
                      & 16                               & 18.04/ 18.37/ 18.82 & 24.78/ 24.64/ 23.94 & 0.0086/ 0.0084/ 0.0072 & 0.0040/ 0.0037/ 0.0035 \\ \hline
\end{tabular}%
}
\caption{Quantitative evaluation of our proposed Bayesian uncertainty reduction technique based adversarial defense mechanism.}
\label{tab: attack_defend}
\end{table*}

\begin{figure}[!ht]
    \centering
    \subfigure[Scale: $\times $2, Adv. Attack: 4]
    {
        \includegraphics[width=4cm]{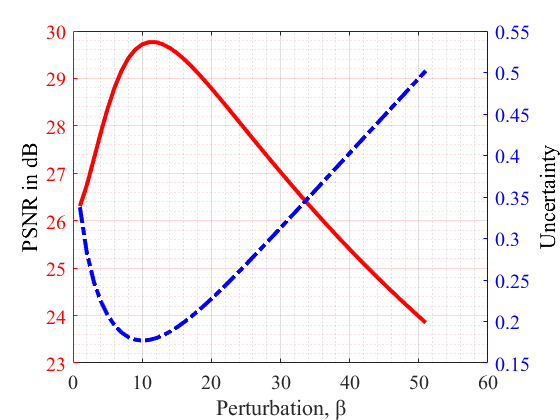}
    } \hspace{-0.2cm}
    \subfigure[Scale: $\times $4, Adv. Attack: 8]
    {
        \includegraphics[width=4cm]{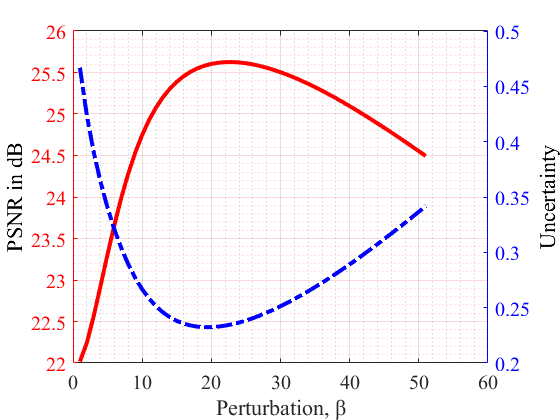}
    } 
\caption{Effect of perturbation level for adversarial defense on quality of the image and mean uncertainty of the corresponding image.}
\label{fig: plots_adv}
\end{figure}

\begin{table*}[ht]
\centering
\resizebox{\textwidth}{!}{%
\begin{tabular}{|c|c|c|c|c|c|}
\hline
\multirow{2}{*}{\begin{tabular}[c]{@{}c@{}}Noise\\ Level\end{tabular}} & \multicolumn{5}{c|}{Scale Factor ($\times $2/ $\times $3/ $\times $4)}                                                              \\ \cline{2-6} 
 & RCAN~\cite{zhang2018image} & SAN~\cite{dai2019second} & DRLN~\cite{anwar2020densely} & BayesianVDSR & \begin{tabular}[c]{@{}c@{}}BayesianVDSR\\ (Uncertainty Reduction)\end{tabular} \\ \hline
5                      & 33.10/ 31.87/ \textbf{30.33} & 33.70/ 31.76/ 30.24 & 33.73/ \textbf{31.94}/ 30.29 & 33.44/ 31.28/ 29.70 & \textbf{34.08}/ 31.69/ 29.78 \\ \hline
10                     & 28.94/ 28.36/ 27.46 & 29.41/ 28.08/ 27.11 & 29.44/ 28.45/ 27.30 & 29.32/ 28.06/ 27.18 & \textbf{31.68}/ \textbf{29.73}/ \textbf{28.17} \\ \hline
15                     & 26.11/ 25.49/ 24.94 & 26.26/ 25.08/ 24.53 & 26.27/ 25.65/ 24.52 & 26.43/ 25.61/ 25.04 & \textbf{29.65}/ \textbf{28.31}/ \textbf{27.12} \\ \hline
\end{tabular}%
}
\caption{Zero-mean Gaussian noise with different variances is added to the low-resolution images of the Set5 dataset. We compare the performance of state-of-the-art models with Bayesian VDSR and its uncertainty reduction counter-part. Note that none of the models is designed to handle the noise. Best results are in bold.}
\label{tab: noise}
\end{table*}

\begin{table*}[]
\centering
\resizebox{\textwidth}{!}{%
\begin{tabular}{|c|c|c|c|c|}
\hline
\multirow{3}{*}{\begin{tabular}[c]{@{}c@{}}Noise \\ Type\end{tabular}} &
  \multicolumn{4}{c|}{Scale Factor ($\times $2/ $\times $3/ $\times $4)} \\ \cline{2-5} 
 & \multicolumn{2}{c|}{PSNR} & \multicolumn{2}{c|}{Uncertainty} \\ \cline{2-5}  
        & Before UN Reduction              & After UN Reduction              & Before UN Reduction                 & After UN Reduction                  \\ \hline
Poisson & 29.16/ 28.04/ 27.10 & 31.28/ 29.49/ 28.10 & 0.0023/ 0.0026/ 0.0026 & 0.0007/ 0.0010/ 0.0013 \\ \hline
Speckle & 27.80/ 26.83/ 26.22 & 29.73/ 28.58/ 27.45 & 0.0025/ 0.0029/ 0.0030 & 0.0010/ 0.0015/ 0.0013 \\ \hline
\end{tabular}%
}
\caption{Performance of our uncertainty reduction technique on noisy LR images.}
\label{tab: othernoise}
\end{table*}

\subsection{Uncertainty Reduction}

\subsubsection{Defense Against Adversarial Attack}

In the earlier section, we show that uncertainty holds a strong relation with different adversarial attack levels. The model's performance degrades severely in the adversarial attacked images, and the corresponding uncertainty map produces more uncertainties compared to non-adversarial images. As mentioned in Algorithm~\ref{algo3}, we propose an uncertainty reduction technique that reduces pixel-wise deviations between MC samples, and we successfully defend the adversarial attack by reducing uncertainties.

In our experiments, we apply the I-FGSM attack~\cite{advattack} on the single image super-resolution model, as shown in ~\cite{choi2019evaluating} and it is also discussed in supplementary. We use the proposed uncertainty reduction based defense mechanism on five different levels of adversarial attacked images. The experimental results are presented in Table~\ref{tab: attack_defend}. We observe from the table that our model reduces uncertainties between MC samples and increase the model's reconstruction performance in terms of PSNR. All the experiments in Table~\ref{tab: attack_defend} are performed on the BSD100 dataset. The table itself depicts that an increase in the level of attacks drastically reduces the reconstruction performance. Our defense mechanism successfully prevented the free fall of reconstructed image quality under adversarial attacks. We also observe that the defense mechanism does not affect non-adversarial images and posses negligible performance drop. The minute drop may be due to slightly blurring the higher uncertainty regions in the non-adversarial images like edges during uncertainty reduction. Figure~\ref{fig: AdvDef} shows a subjective evaluation of our proposed uncertainty reduction based defense mechanism. We observe that low-resolution (LR) and adversarial LR images do not significantly differ in perception. However, the $\times 2$ scaled super-resolved (SR) adversarial image without defense exhibits many artifacts in the image and produces more uncertainty, as shown in the uncertainty (UN) map. Our proposed defense mechanism significantly reduces those unwanted artifacts, and uncertainty also reduces.

In our uncertainty reduction technique, we perturb a LR test image during uncertainty reduction so that uncertainty reduces for that test image. The perturbation level $\beta$ plays an essential role. As we increase the perturbation level, the uncertainty reduces, and reconstructed image quality increases, as shown in Figure~\ref{fig: plots_adv}. After the optimum perturbation level, the uncertainty starts to increase due to excessive perturbation. The optimum perturbation level depends on the attack level and scale factor model we use. After getting gradient directions, we choose the perturbation level, which produces minimum uncertainty during reconstruction.

\subsubsection{Noisy Low-resolution Image}

Most of the deep super-resolution models handle bicubic downsampled images as input and is trained to map high-resolution counterparts of inputs. However,  the real-world LR images are not the ideal bicubic downsampled version of high-resolution images. Noise is one of the critical degradation factors incorporated with real-world LR images. Some approaches consider that noise in the form of Gaussian noise and train the network. It performs well on the images that follow the Gaussian distribution. However, there is no guarantee that it will work on real-world images as real-world noise distribution varies based on the camera sensor and environmental effects. In the earlier section, we have observed that small noise degrades the Bayesian VDSR network's performance and increases the uncertainty. Therefore, we apply the proposed uncertainty reduction technique to reduce the uncertainty induced due to noisy LR images. Our experimental findings in Table~\ref{tab: noise} show that the uncertainty reduction technique improves the quality of the SR images from noisy LR counterparts where noise follows Gaussian distribution. As our model is not trained on noisy LR images, therefore for a fair comparison, we compare state-of-the-art deep learning models trained on bicubic LR images. We observe from the table that all the models' performance degrades due to the increase of noise, and our proposed uncertainty reduction technique can prevent degradation to some extent. In our experiment, we perform a single update on the LR image for zero-mean Gaussian noise with variance 5, and we update the LR image $10$ times for noise variance $10$ and $15$. We also experiment with other different types of noisy LR images. We consider Poisson and Speckle noise in the LR images and the uncertainty reduction performance is shown in Table~\ref{tab: othernoise}. We witness a consistent improvement of image quality in those noisy LR images due to uncertainty reduction.

\section{Conclusion}
\label{sec:conclude}
In this work, we introduced a  faster Bayesian approach to estimate uncertainty in batch normalized super-resolution network. We experimentally found that the uncertainty map is like the signature map of the SISR model, which indicates how good is the model in reconstructing images. We experimentally found the small adversarial noises and even random Gaussian noises on LR images increase the uncertainty as those images do not follow bicubic downsampled training image distribution. It suggests that uncertainty represents the limitation of the model. We propose an uncertainty reduction method between MC samples by perturbing the test LR image, and it eventually prevents the adversarial attack. We have also shown that the uncertainty reduction technique is a useful tool for reconstruction performance improvement on noisy LR images. We believe that this work will set a important gateway into the study of Bayesian deep learning. A more efficient way to reduce uncertainty will make the neural network work admirably for out-of-distribution samples, and it will benefit in handling unknown degradations. 

\begin{center}
 {\larger[4] \color{red}Supplementary Material\newline}
\end{center}

\setcounter{section}{0}
\setcounter{figure}{0}
\setcounter{table}{0}

\section{Adversarial Attack Mechanism}

The modified version of the Iterative Fast Gradient Sign Method (I-FGSM) is used to perform the adversarial attack on the super-resolution model~\cite{choi2019evaluating}. The main goal of the adversarial attack on the super-resolution model is to perturb the low-resolution (LR) image so that the super-resolution model fails to perform and creates unwanted artifacts. Following the paper, we use two types of attacks: Basic Attack and Partial Attack. We adversarially perturb the whole LR image in the Basic Attack and a small portion of the LR image for Partial Attack.

\textbf{Basic Attack:} Let $f(\cdot)$ is the super-resolution model. $I_{L_0}$ is the LR image, and $I_L$ is the adversarially perturbed LR image. $f(I_{L_0})$ and $f(I_L)$ are the super-resolved images. The main objective is to maximize the difference between $f(I_{L_0})$ and $f(I_L)$. It can be defined as:
\begin{equation}
    \mathcal L(I_L, I_{L_0})=\parallel f(I_L) - f(I_{L_0}) \parallel_{2}
\end{equation}
$I_L$ can be iteratively updated by using the I-FGSM update rule by:
\begin{equation}
    \resizebox{.9\hsize}{!}{$\widetilde{I}_L(N+1)=clip_{0,1}(I_L(N)+\frac{\alpha}{T}sign(\triangledown \mathcal L(I_L(N), I_{L_{0}})))$}
\end{equation}
\begin{equation}
    I_L(N+1)=clip_{-\alpha, \alpha}(\widetilde{I}_L(N+1)-I_{L_{0}})+I_{L_{0}}
\end{equation}
Here, $T$ is the number of iterations, $sign(\triangledown \mathcal L(I_L(N), I_{L_{0}}))$ is the sign of the gradient, $clip_{0,1}$ clip the pixel values in between $0$ and $1$, and $\alpha$ is the adversarial noise level.

\textbf{Partial Attack} In case of a partial attack, we restrict the attack region and add adversarial noise in that specific region. Let $M$ is the masked region, where the area that will be attacked is set to $1$, and the rest is $0$. Therefore, the new I-FGSM update rule is defined as,
\begin{equation}
    \resizebox{.9\hsize}{!}{$\widetilde{I}_L(N+1)=clip_{0,1}(I_L(N)+\frac{\alpha}{T}sign(\triangledown \mathcal L(I_L(N), I_{L_{0}}))\circ M)$}
\end{equation}
where $\circ$ is the pixel-wise multiplication.

\begin{figure}[!ht]
    \centering
    \subfigure[PSNR vs Uncertainty (UN)]
    {
        \includegraphics[width=6cm]{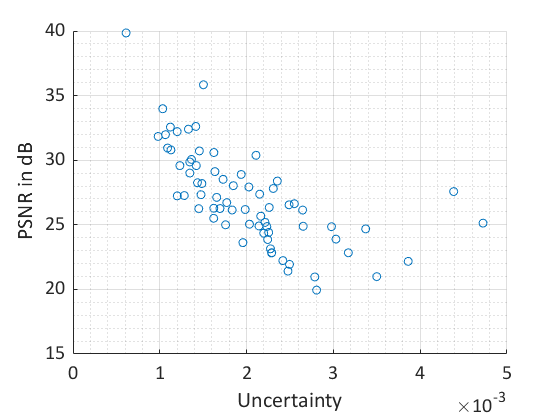}
        \label{fig: plots_PvsUN}
    } \hspace{-0.2cm} 
    \subfigure[UN vs Scale]
    {
        \includegraphics[width=4cm]{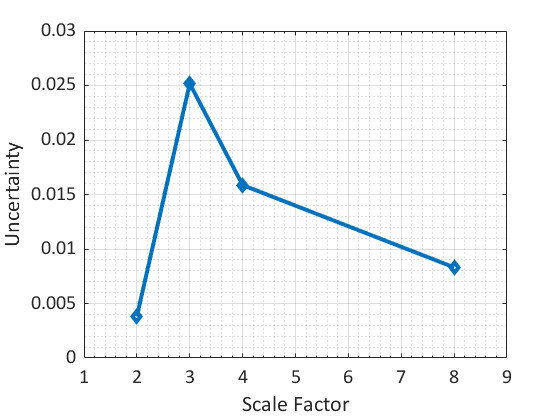}
        \label{fig: plots_UNvsS}
    } \hspace{-0.2cm} 
    \subfigure[UN vs Adv. Noise Level]
    {
        \includegraphics[width=4cm]{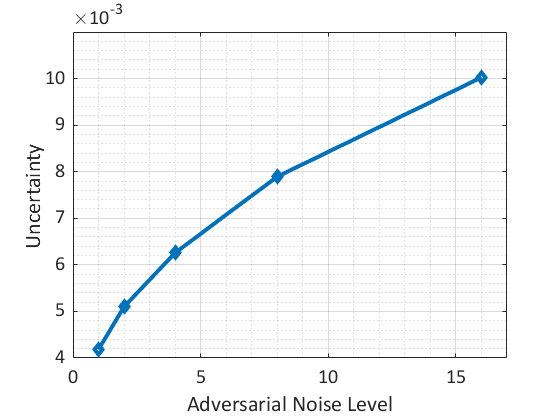}
        \label{fig: plots_UNvsA}
    }
\caption{(a) presents the relation between PSNR and Bayesian uncertainty. (b) presents the influence of uncertainty for different scale factor models. (c) presents impact of adversarial noise levels with uncertainty. (zoom for the best view.)}
\label{fig: plotsx}
\end{figure}

\begin{table*}[!ht]
\centering
\resizebox{\textwidth}{!}{%
\begin{tabular}{|c|c|c|c|c|c|}
\hline
\multirow{2}{*}{Dataset} &
  \multirow{2}{*}{Attack Level} &
  \multicolumn{2}{c|}{\begin{tabular}[c]{@{}c@{}}Uncertainty\\ Scale Factor $\times 3$/ $\times 4$/ $\times 8$\end{tabular}} &
  \multicolumn{2}{c|}{\begin{tabular}[c]{@{}c@{}}PSNR\\ Scale Factor $\times 3$/ $\times 4$/ $\times 8$\end{tabular}} \\ \cline{3-6} 
                          &           & No Defense             & After Defense          & No Defense          & After Defense       \\ \hline
\multirow{6}{*}{BSDS100}  & No Attack & 0.0252/ 0.0159/ 0.0083 & 0.0019/ 0.0016/ 0.0018 & 27.57/ 26.61/ 24.56 & 27.88/ 27.23/ 24.56 \\ \cline{2-6} 
                          & 1         & 0.0225/ 0.0153/ 0.0075 & 0.0026/ 0.0019/ 0.0021 & 26.04/ 25.53/ 23.80 & 27.60/ 26.96/ 24.37 \\ \cline{2-6} 
                          & 2         & 0.0197/ 0.0146/ 0.0075 & 0.0028/ 0.0023/ 0.0024 & 23.78/ 24.08/ 22.66 & 27.45/ 26.70/ 24.28 \\ \cline{2-6} 
                          & 4         & 0.0282/ 0.0202/ 0.0103 & 0.0031/ 0.0027/ 0.0029 & 20.27/ 21.25/ 20.50 & 26.88/ 26.24/ 24.07 \\ \cline{2-6} 
                          & 8         & 0.0270/ 0.0217/ 0.0121 & 0.0036/ 0.0033/ 0.0035 & 17.25/ 18.21/ 18.03 & 26.14/ 25.50/ 23.66 \\ \cline{2-6} 
                          & 16        & 0.0308/ 0.0290/ 0.0156 & 0.0048/ 0.0045/ 0.0045 & 14.97/ 15.28/ 15.50 & 24.77/ 24.17/ 22.85 \\ \hline
\multirow{6}{*}{Urban100} & No Attack & 0.0257/ 0.0325/ 0.0119 & 0.0031/ 0.0030/ 0.0034 & 26.21/ 24.17/ 22.02 & 25.95/ 25.03/ 21.94 \\ \cline{2-6} 
                          & 1         & 0.0276/ 0.0344/ 0.0168 & 0.0038/ 0.0032/ 0.0038 & 24.66/ 23.02/ 20.94 & 25.54/ 24.68/ 21.63 \\ \cline{2-6} 
                          & 2         & 0.0291/ 0.0369/ 0.0156 & 0.0039/ 0.0036/ 0.0042 & 22.67/ 21.51/ 19.82 & 25.34/ 24.17/ 21.45 \\ \cline{2-6} 
                          & 4         & 0.0331/ 0.0392/ 0.0201 & 0.0040/ 0.0038/ 0.0047 & 19.71/ 19.26/ 17.98 & 24.93/ 23.99/ 21.23 \\ \cline{2-6} 
                          & 8         & 0.0303/ 0.0503/ 0.0192 & 0.0043/ 0.0046/ 0.0051 & 16.73/ 16.69/ 16.00 & 24.41/ 23.27/ 20.93 \\ \cline{2-6} 
                          & 16        & 0.0320/ 0.0523/ 0.0222 & 0.0056/ 0.0057/ 0.0061 & 14.31/ 14.20/ 14.02 & 23.24/ 22.25/ 20.37 \\ \hline
\multirow{6}{*}{Manga109} & No Attack & 0.0170/ 0.0123/ 0.0081 & 0.0024/ 0.0025/ 0.0030 & 31.26/ 28.89/ 24.02 & 31.50/ 29.29/ 23.78 \\ \cline{2-6} 
                          & 1         & 0.0142/ 0.0118/ 0.0085 & 0.0030/ 0.0028/ 0.0032 & 29.44/ 27.68/ 23.39 & 30.59/ 28.87/ 23.59 \\ \cline{2-6} 
                          & 2         & 0.0165/ 0.0143/ 0.0104 & 0.0036/ 0.0033/ 0.0034 & 26.85/ 25.87/ 22.41 & 29.36/ 27.88/ 23.26 \\ \cline{2-6} 
                          & 4         & 0.0258/ 0.0160/ 0.0097 & 0.0044/ 0.0038/ 0.0039 & 22.94/ 22.94/ 20.80 & 28.15/ 26.89/ 22.84 \\ \cline{2-6} 
                          & 8         & 0.0294/ 0.0185/ 0.0124 & 0.0051/ 0.0046/ 0.0046 & 19.09/ 19.50/ 18.47 & 26.77/ 25.71/ 22.24 \\ \cline{2-6} 
                          & 16        & 0.0267/ 0.0285/ 0.0147 & 0.0059/ 0.0058/ 0.0057 & 16.11/ 16.22/ 15.95 & 24.88/ 23.89/ 21.31 \\ \hline
\end{tabular}%
}
\caption{Quantitative evaluation of our proposed Bayesian uncertainty reduction technique based adversarial defense mechanism.}
\label{tab: my-table}
\end{table*}

\section{Additional Experiments: Residual Network}
We perform similar experiments on another deep neural architecture like VDSR, as described in the main paper. For this purpose, we use the popular SR-ResNet architecture~\cite{ledig2017photo}, which consists of consecutive residual blocks.

\subsection{Network Architecture}

The residual network is one of the most popular architecture for image classification. SR-ResNet~\cite{ledig2017photo} uses the consecutive residual blocks for super-resolution. We adopt the SR-ResNet architecture with some minor modifications. In our architecture, the residual block consists of two consecutive convolution layers and each followed by ReLU and batch-normalization. There are 16 consecutive residual blocks in the network.

\subsection{Training Details}

We use the DIV2K dataset~\cite{dataset, datasetpaper} for training, which contains $800$ training images and $100$ images for validation. Five standard benchmark testing datasets, namely Set5~\cite{set5}, Set14~\cite{set14}, BSD100~\cite{bsd100}, Urban100~\cite{urban100}, Manga109~\cite{manga109} are used for performance analysis. We randomly extract patches of size $32\times 32$ from each LR image during training for a batch update. Each batch contains $16$ patches. We augment the patches by horizontal flip, vertical flip, and $90$-degree rotation and randomly choose each augmentation with a $50\%$ probability. Each input patch is normalized into $[0, 1]$ and $[0.4488, 0.4371, 0.4040]$ is subtracted channel-wise before feeding to the network. We train each model with the PyTorch framework for $1000$ epochs, where a single epoch constitutes $1000$ batch updates. Adam optimizer~\cite{adam} is used to update the weights. The learning rate is initialized to $10^{-4}$ and reduced to half after every $200$ iterations. We use mean-squared error to optimize model parameters.

\subsection{Behaviour of Uncertainty}

Like VDSR architecture~\cite{vdsr}, as described in the main paper, the Figure~\ref{fig: plots_PvsUN} shows a strong relationship between reconstructed image quality and uncertainty in reconstruction. BSDS100 dataset is used for this experiment, and the experiment is performed for scale factor $\times4$. We observe that images with lower PSNR, which represents the image quality, give higher uncertainty. Figure~\ref{fig: plots_UNvsS} shows the relation between scale factor and uncertainty. Unlike VDSR, we do not observe any relationship between those. In Figure~\ref{fig: plots_UNvsA}, We witness that images with more adversarial perturbation lead to higher uncertainty and experiment is performed for scale factor $\times 2$.

\subsection{Defense Against Adversarial Attack}

The performance of our proposed adversarial defense mechanism on the SR-ResNet model is shown in Table~\ref{tab: my-table} using the three most popular testing datasets, namely BSDS100, Urban100, and Manga109. Like the main paper, we use the five different adversarial attack levels, as shown in ~\cite{choi2019evaluating}, to show the efficacy of the defense mechanism. We also show the performance of the defense mechanism on the images which are not perturbed adversarially. The performances are shown for three different scale factors, that is, $\times3$, $\times4$, $\times8$.

Like VDSR architecture in the main paper, Table~\ref{tab: my-table} shows similar performance in preventing adversarial attacks while we use SR-ResNet architecture and several datasets like BSDS100, Manga109, and Urban100. Our proposed defense mechanism performs well to improve the reconstruction performance in adversarially attacked images. As the attack level increases, the performance drops drastically. Our proposed method successfully prevents that performance drop.

{\small
\bibliographystyle{ieee_fullname}
\bibliography{egbib}
}

\end{document}